\documentclass[runningheads]{llncs}
\usepackage{graphicx}
\usepackage{amsfonts}
\usepackage{algorithm}
\usepackage{algorithmic}

\begin{document}
\title{Achieving Transparency in Distributed Machine Learning with Explainable Data Collaboration}
\titlerunning{Explainable Data Collaboration}
%

\author{Anna Bogdanova\inst{1,2} \and
Akira Imakura\inst{1,2} \and
Tetsuya Sakurai\inst{1,2} \and 
Tomoya Fujii\inst{3} \and Teppei Sakamoto\inst{3} \and Hiroyuki Abe\inst{3}}

\authorrunning{A. Bogdanova et al.}
%
\institute{Center for Artificial Intelligence Research, University of Tsukuba, Japan \and
Department of Computer Science, University of Tsukuba, Japan
\and
NTT DATA Corporation, Japan\\}

\authorrunning{A. Bogdanova et al.}
%
%
\maketitle              
\begin{abstract}
Transparency of Machine Learning models used for decision support in various industries becomes essential for ensuring their ethical use. To that end, feature attribution methods such as SHAP (SHapley Additive exPlanations) are widely used to explain the predictions of black-box machine learning models to customers and developers. However, a parallel trend has been to train machine learning models in collaboration with other data holders without accessing their data. Such models, trained over horizontally or vertically partitioned data, present a challenge for explainable AI because the explaining party may have a biased view of background data or a partial view of the feature space. As a result, explanations obtained from different participants of distributed machine learning might not be consistent with one another, undermining trust in the product. This paper presents an Explainable Data Collaboration Framework based on a model-agnostic additive feature attribution algorithm (KernelSHAP) and Data Collaboration method of privacy-preserving distributed machine learning. In particular, we present three algorithms for different scenarios of explainability in Data Collaboration and verify their consistency with experiments on open-access datasets. Our results demonstrated a significant (by at least a factor of 1.75) decrease in feature attribution discrepancies among the users of distributed machine learning.

\end{abstract}

\section{Introduction}
\label{S:1}


With more actors in the industry and public sector adopting AI-based solutions to enhance their services, their impact on society and individuals comes under scrutiny from ethical and safety concerns. A systematic review of existing principles and guidelines for ethical AI \cite{jobin2019global} over 84 documents from 12 countries showed transparency as the most agreed-upon ethical AI principle, featured in 73 out of 84 sources. Specifically, in the context of algorithmic transparency, there is a demand for human-readable explanations of algorithmic decisions so that they could be appealed, and their fairness attested \cite{zerilli2019transparency}. The task of interpreting and explaining decisions of complex machine learning models in a human-readable form became the focus of an active field of research on Explainable AI.

Another aspect of ethical use of AI concerns the responsible handling of personal data. Many countries have legal regulations, such as the EU GDPR and CCPA, limiting the use and transfer of users' data. Current solutions to privacy-preserving AI rely upon the techniques of distributed machine learning, encryption, and data perturbation in order to decouple the learned model from training data \cite{chen2020developing} \cite{cheng2020federated}, \cite{imakura2020data}. Usually, multiple users would engage in a collaborative training of a machine learning model without exposing their private data, each gaining a trained model as a result of collaboration. 

The decoupling of the learned model from the training data naturally presents challenges for explainability \cite{grant2020show}. To some extent, a workaround is offered by model-agnostic explainability methods, such as Shapley Additive Explanations (SHAP). It allows treating the collaboratively trained model as a black-box, probe it with various inputs, and assign relative feature importance attributions from the analysis of the outputs. In addition, SHAP method is known for the intuitive presentation of explanation results and the ability to expose bias in training data \cite{jain2020biased}. However, because the explanations obtained in such a manner are relative to the background datasets coming from different distributions, the explanations might not be consistent among the data-holders.

This presents several challenges for transparency of machine learning models trained in a distributed setting: 
i) if some users are members in more than one horizontally partitioned dataset, they should not get contradictory explanations for the same model from different data holders;
ii) if data bias present in one data holder has been corrected through collaboration, the obtained explanations should no longer display the bias; 
iii) if users have their data vertically partitioned among several data holders, they should be able to obtain correctly proportioned additive feature attributions for the complete set of features. (e.g., Shapley values for all features added to the base value should equal the actual model prediction);
iv) data holders in vertical data collaboration should be able to inspect 'host' feature attributions with 'guest' features hidden; simultaneously, their view of feature attributions should be consistent with the user-side view for the complete set of features.


As there are many varieties to distributed machine learning, each of them might need a separate approach to model explainability. In this paper, we focus on one method of distributed privacy-preserving machine learning - Data Collaboration Analysis. This method presents an additional challenge to explainability because it involves individual data transformations on the feature space. As a result, users who are executing explainability algorithms might have different sample distributions, different sets of features, and partially different model components. Our goal is to adapt SHAP explainability to Data Collaboration setting in a way that provides consistent explanations for each user at the same time not revealing information about the features that are not shared.   


The main contributions of this work are summarized as follows:
\begin{enumerate}
    \item We identify the problem with transparency that the conventional application of the SHAP method in distributed machine learning may produce and demonstrate it with a case study on open-access data.  
    \item We propose a Horizontal DC-SHAP algorithm that uses a shareable anchor dataset as a baseline to produce consistent explanations among all collaborators and experimentally verify the consistency of our method on open data.
    \item We propose a Vertical DC-SHAP(i) algorithm that provides client-side feature attributions for the whole set of features and a Vertical DC-SHAP(ii) algorithm that provides a partial view of feature attributions visible to one of the data holders; we further demonstrate the consistency of algorithms (i) and (ii).
\end{enumerate}

The problem of explainability in Federated Learning has been previously addressed from the perspective of maintaining the privacy of guest features in vertically partitioned datasets \cite{wang2019interpret}. However, the problem of inconsistency of explainability algorithms in distributed machine learning, to the best of our knowledge, is being addressed for the first time.



\section{Related Work}
\label{S:2}

\subsection{Distributed Machine Learning}

Distributed and privacy-preserving machine learning has been explored in a wide range of approaches, with the common goal of learning a model without exposing training data to the analyst or all possible users of such model. Most approaches involve some way of aggregation of model parameters. For instance, in Private Aggregation of Teacher Ensembles (PATE) \cite{papernot2018scalable} privately trained models are used to train a global student model with differential privacy. Another popular approach is Federated Learning that trains a global machine learning model through iterative rounds of averaging of local models' parameter updates \cite{konevcny2016federated}. In addition, Vertical Federated Learning is concerned with a particular problem setting where data is partitioned in feature space rather than in sample space \cite{yang2019federated}. 




\subsection{Data Collaboration}
\label{S:2-2}

\begin{figure}
\centering
\includegraphics[width=1\linewidth]{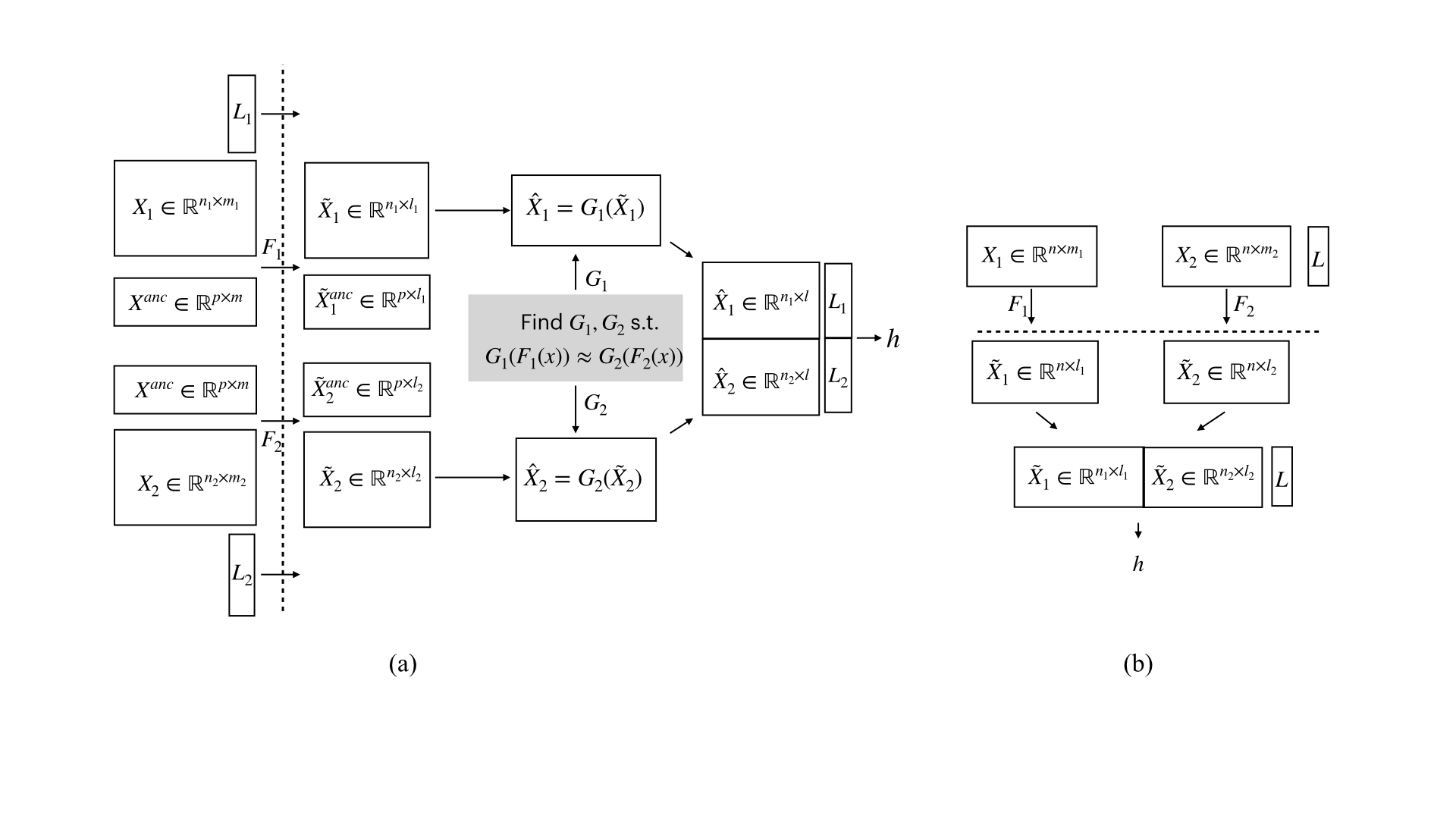}
\caption{Horizontal DC (a) and Vertical DC (b), where $X_i$, are private datasets, $f_i$ are private transformation functions, $X^{anc}$ is shared anchor data, $L_i$ are labels for supervised learning tasks, $g_i$ are integrating transformation functions, and $h$ is a machine learning model\label{fig:dc1}}

\end{figure}

The Data Collaboration (DC) method was proposed in 2020 \cite{imakura2020data}. Different from Federated Learning, it is a non-model-sharing type of distributed machine learning, where participants have different model components in their pipeline \cite{bogdanova2020federated}, \cite{imakura2021collaborative}. It employs irreversible transformations of dimensionality reduction that are executed by users locally to produce shareable intermediate representations of their data. The central analyzer then integrates these intermediate representations into one dataset and trains a machine learning model, which is further distributed back to each user. The integration step is different for horizontal and vertical partition settings. While in a vertical partition setting, integration consists of a simple concatenation operation, in a horizontal partition setting, additional integrating transformation is necessary. This transformation is computed with the help of auxiliary synthetic data shared among the users. In the original Data Collaboration paper, such data is called anchor data, and it is produced by collaborators as randomly generated values within the distribution of original data features. Figure \ref{fig:dc1} outlines the horizontal and vertical Data Collaboration mechanisms for two users. The task of interpretability in Data Collaboration, the obtaining of overall feature importance, has been previously developed through building a surrogate model from DC model predictions on shareable anchor data \cite{imakura2021interpretable}. The focus of this paper is explainability, which is concerned with feature importance attribution for individual samples.

\subsection{Explainable Machine Learning}

The goal of explainability is to interpret the prediction given by a model on a given input by attributing relative importance to each feature of the input. Some models (e.g. linear regression) naturally offer such explainability, therefore they are regarded as interpretable machine learning models. By fitting a line to the data, we obtain weights of input features that add up to the prediction. However, in most machine learning models, some interpretable approximation to the original model must be used. The two commonly used methods of model-agnostic explainability are LIME \cite{ribeiro2016model} and SHAP \cite{lundberg2017unified}. The latter work proposed that several explainability methods can be unified under the formulation of additive feature attribution methods,

\begin{equation}
    g(z') = \phi_0 + \sum^M_{i=1}{\phi_iz'_i}
\end{equation}

where $z' \in{\{0,1\}}^M$ is a binary variable representing the presence or absence of simplified input features $M$ and $\phi_i\in{\mathbb{R}}$ is an attribution of feature importance. 

In the same paper, the authors conjectured that only the methods based on game-theoretic Shapley values satisfy three desirable properties of model explanations: accuracy, missingness, and consistency. For the purposes of this paper, the consistency property is of special importance. It states that for any two models $f$ and $f'$, if 

\begin{equation}
\label{eq_consistency}
    f'_x(z') - f'_x(z' / i) \ge f_x(z') - f_x(z' / i)
\end{equation}

for all inputs $z' \in{\{0,1\}^M}$, where $z'/i$ denotes instances of $z'$ where $z'_i = 0$, then $\phi_i(f', x) \ge \phi_i(f,x)
$. In other words, for a given sample $x$, if the change in the output of $f'$ from introduction of a simplified feature $z'_i$  is bigger or equal to the change in the output of $f$, the attribution value for the feature $i$ in model $f'$ should not be less than its attribution in the model $f$.  

To extend the consistency principle to Data Collaboration method, we let $f$ be the collaboration model $h$ composed with individual transformation functions $f_1$ and $g_1$ of the first user, and $f'$ be the collaboration model $h$ composed with the functions $f_2$ and $g_2$ of the second user (Figure~\ref{fig:dc1}(a)). Then, according to the inequality (2), if the collaboration model of one user relies more on a certain feature, this feature should get bigger attribution in the explanations for this user. 
This important property makes Shapley values-based methods suitable to provide explainability for the Data Collaboration method. 

Algorithm \ref{appendix:shap} in Appendix presents a procedure for a model-agnostic KernelSHAP method \cite{lundberg2017unified}. 



\section{Methodology}
\label{S:3}

\subsection{Feature Attribution in Horizontal Data Collaboration}


In horizontal collaboration (see Figure~\ref{fig:dc1} (a)), participants share the same set of features, but non-iid distribution of data samples can prevent getting consistent explanations. It happens because the available background dataset determines the reference value in the calculations of SHAP values. 

We solve the issue of different background datasets by producing the reference value from the anchor data which is already shared among the users. In particular, we take the median value of the anchor data, though other aggregate statistics are also possible.

In our method (see Algorithm \ref{appendix:hor} in Appendix), a user who wishes to get feature attributions for a model prediction runs SHAP algorithm on the inputs of the sample of interest $x$, reference value $r$ calculated from the anchor data, and a  model composed of transformation functions $F$ and $G$ individual to the user, and a shared machine learning model $h$. In this way, the only difference from the conventional KernelSHAP algorithm consists in the composition of the model under explanation.  However, it is going to be the case that composed models of different users may give different attributions to the features of the same sample. We claim that since SHAP values satisfy the principle of consistency, the discrepancies in feature attributions among the users truthfully reflect the discrepancies in model components trained in collaboration.

\begin{figure}
\centering
\includegraphics[width=0.6\linewidth]{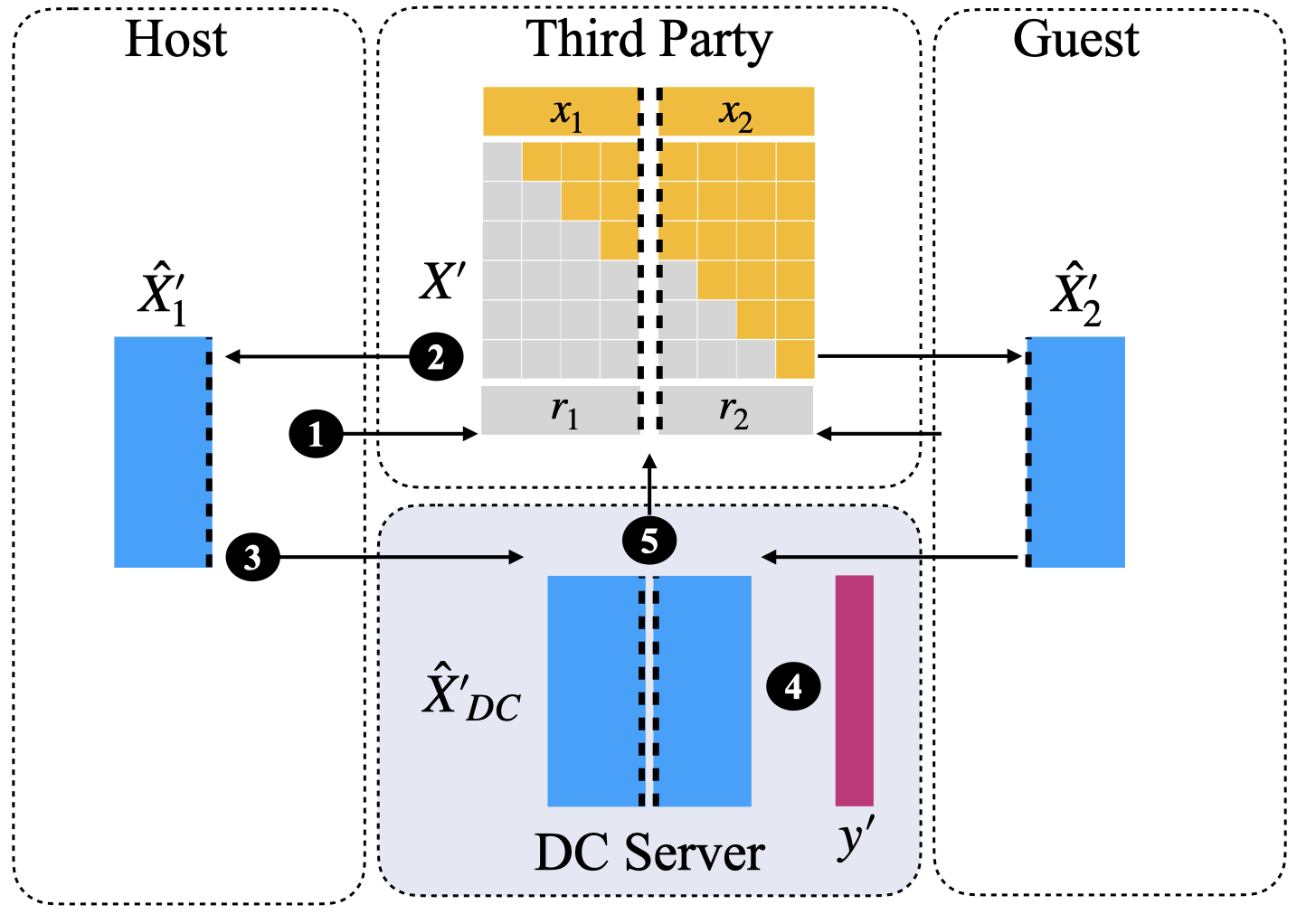}
\caption{Calculating SHAP values for all features in vertical Data Collaboration. 1) Third-party user gets reference values (can be chosen by the third party or supplied by collaborators); 2) third-party user constructs artificial input data as a powerset of reference and sample values; 3) collaborators transform corresponding partial input data; 4) collaborators centralize input data and get model prediction; 5) third-party user computes SHAP values with Algorithm \ref{appendix:shap} \label{fig:shap_dc1}}
\end{figure}

\begin{figure}

\centering\includegraphics[width=0.6\linewidth]{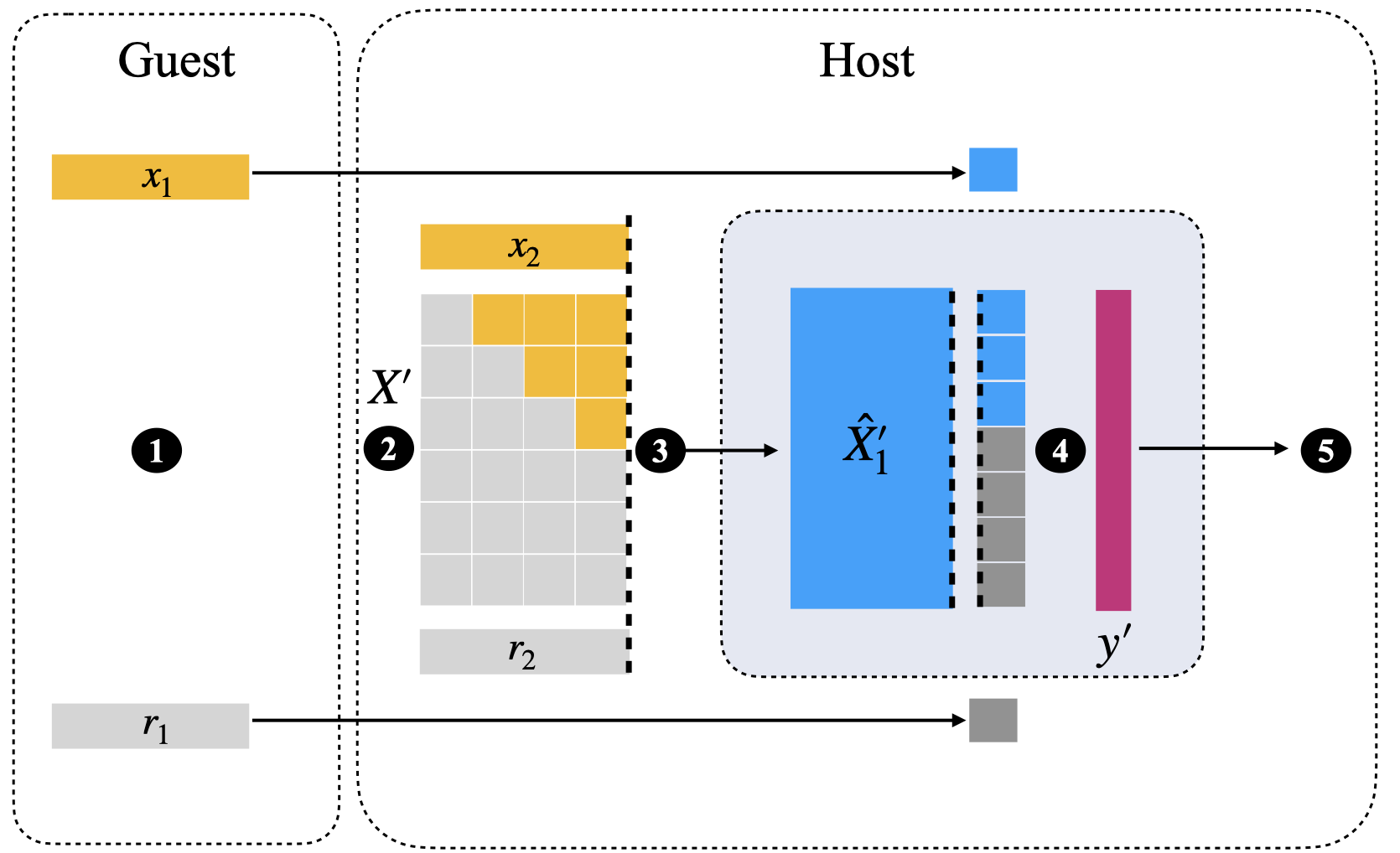}
\caption{Computing partial SHAP values in vertical DC. 1) Guest party supplies the intermediate representations of the missing sample and reference values; 2) host party constructs partial input data as a powerset of sample and reference feature values; 3) host party gets intermediate representation of the partial input data; 4) host party constructs an aggregated features vector with intermediate representations obtained from the guest; 5) collaborators unify intermediate representations and get model predictions; 6) Host party computes SHAP values with Algorithm \ref{appendix:shap} in Appendix)\label{fig:shap_dc2}}
\end{figure}

\subsection{Feature Attribution in Vertical Data Collaboration}

In vertical collaborative (see Figure~\ref{fig:dc1} (b)), users have overlapping data points but different sets of features. Such situation often arises in a business setting when separate entities collect different information on the same set of individual clients.

The challenge for explainability arises because the feature sets held by the users are disjoint, and every evaluation of a sample by the model requires a contribution of the missing features by all users. For simplicity, we will assume the collaboration of two users, whom we call the host and the guest, following the convention in literature. Hence, if host wishes to evaluate the model on a new data point $x$, they must obtain the intermediate representation of the missing features of $x$ held by the guest.  

We imagine two distinct use cases for the explainability of Data Collaboration in a vertical setting. 
\begin{enumerate}
    \item Attribution is requested at a third party for the complete set of features. 
    \item Attribution is requested by one of the users for the partial set of features. 
\end{enumerate}

The first use case is important, because individuals can often access their own data, and should be able to request explanation of a specific decision made by a collaborative machine learning model for them. Algorithm \ref{appendix:vert_all} (see Appendix) and Figure~\ref{fig:shap_dc1} describe an approach that can produce such explanation. 

In the second use case, when explanations are requested by the host party for the partial set of the features, it is reasonable to maintain secrecy of feature attributions belonging to the guest party. However, displaying the aggregated attributions of all guest features will help to maintain the correct proportions of the host feature impacts on the model. The same approach had been explored by Wang for Federated Learning explainability \cite{wang2019interpret}. 

To achieve such output of feature attribution, in Algorithm \ref{appendix:vert_part} (see Appendix) we construct a power set $S$ of binary feature indicators following the KernelSHAP method (Algorithm \ref{appendix:shap} but with an additional indicator for the aggregated host feature. We then proceed with constructing the simulated model inputs $X'$ consisting of two parts: one for the host features $X'_h$, another for the intermediate representations of the guest features $X'_g$. The host features consequently undergo transformation with host's dimensionality reduction function $F_h$ before both parts of simulated inputs are concatenated and model predictions are obtained. The proposed method is schematically presented in Figure~\ref{fig:shap_dc2}. 

\section{Experiments}
\label{S:4}
\subsection{Experimental Setting}

The experiments presented in this section are set up to validate the proposed algorithms for calculating SHAP values in different Data Collaboration settings. In particular, we are expecting to observe the resemblance of feature attributions for the same inputs when done by different collaborators. However, we do not expect to see the exact equality of feature attributions across the partitions because client-side data transformations necessary for Data Collaboration might influence feature attribution by each client. Such difference in feature attributions is nevertheless correctly reflecting model behavior of the client as follows from the consistency property of SHAP algorithm presented in Section \ref{S:2}. 

Experiments were conducted on Census Income Dataset  (Adult) from UCI Machine Learning Repository \cite{kohavi1996scaling}. It consists of 48842 records of American adults extracted from the 1994 Census database. The prediction task is to determine whether an individual makes over 50K a year. 

The prepossessing steps consisted of encoding all categorical values with labels and dropping features \textit{fnlwgt} and \textit{education} as not informative, shuffling, and separating train and test data. For the model, we used a k-Nearest Neighbours classifier with a number of neighbors set to 7 and "kd-tree" as a solving algorithm. Other Data Collaboration parameters included using 9-dimensional intermediate representations (F) and the same number of dimensions for collaboration projection (G), and 2000 points of anchor data. These parameters were fine-tuned on Adult data to improve the accuracy of the DC model. Unless stated otherwise, the random seed was set to zero.

The experiments were performed on MacBook Pro with a processor of 2.3 GHz 8-Core Intel Core i9 with Python 3.7 development environment.

\subsection{Demonstration of Contradictory Explanations in Distributed Machine Learning}
 The purpose of this experiment is to verify the claim that the application of SHAP explainability independently by each party in distributed machine learning may result in contradictory outputs. 
 
 To demonstrate this, we split the training data so that User 1 holds 90\% of positive labels, and the second user holds the remaining 10\%, setting aside randomly selected 100 samples for validation. As a result, User 1 has 7811 samples with 90\% of positive labels, and User 2 has 24650 samples with only 3\% of positive labels. This produced a biased dataset in possession of User 1  with a higher expected prediction value compared to User 2, which would be reflected in SHAP output for the two users. After the data was split, we trained a Data Collaboration model among the two simulated users and used it to predict the \textit{Income} variable. Next, KernelSHAP method (Algorithm \ref{appendix:shap}) was applied to the DC model and validation data taking User 1 training data as a baseline, then the proposed Horizontal DC-SHAP method (Algorithm \ref{appendix:hor}) was applied, using the shared anchor data as a baseline. The same process was repeated for User 2.
 
 \begin{figure}
\centering\includegraphics[width=0.8\linewidth]{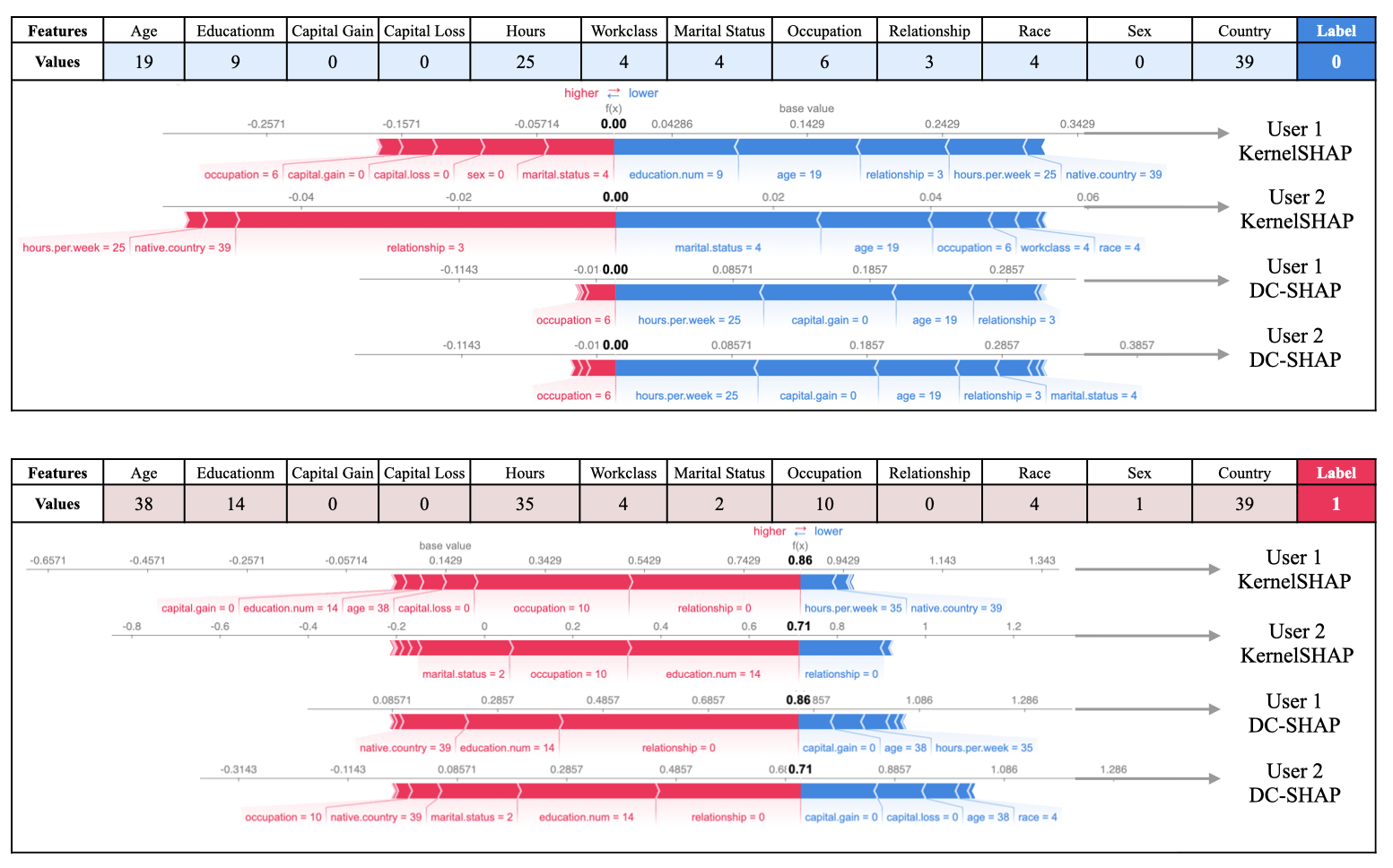}
\caption{Selected samples that demonstrate contradictory explanations. In all cases, the same DC model prediction is being explained for the same data instance. Explanations are given for class 1 predictions. \label{fig:contr}}
\end{figure}

 Figure \ref{fig:contr} shows two samples selected from the validation set that display contradictory explanations when KernelSHAP algorithm is applied.  
 For instance, in Figure \ref{fig:contr} (top), the feature attribution by User 1 of "Marital Status" is positive, while by User 2, it is by an equal amount negative. In Figure \ref{fig:contr} (bottom) it can be observed that the 'Relationship' feature has a positive attribution by User 1 and negative by User 2. By applying DC-SHAP algorithm, the contradictions were resolved.
 It should be noted that the presented cases were hand-picked, and the setting was engineered for the worst-case scenario for demonstration purposes. It is important to remember that in a real-life scenario of distributed machine learning, there will be no way to compare the explanations. Therefore, extreme cases like this should be carefully explored. In the next section, we will verify the consistency of the proposed algorithm in  general cases.

 \begin{figure}

\centering\includegraphics[width=0.7\linewidth]{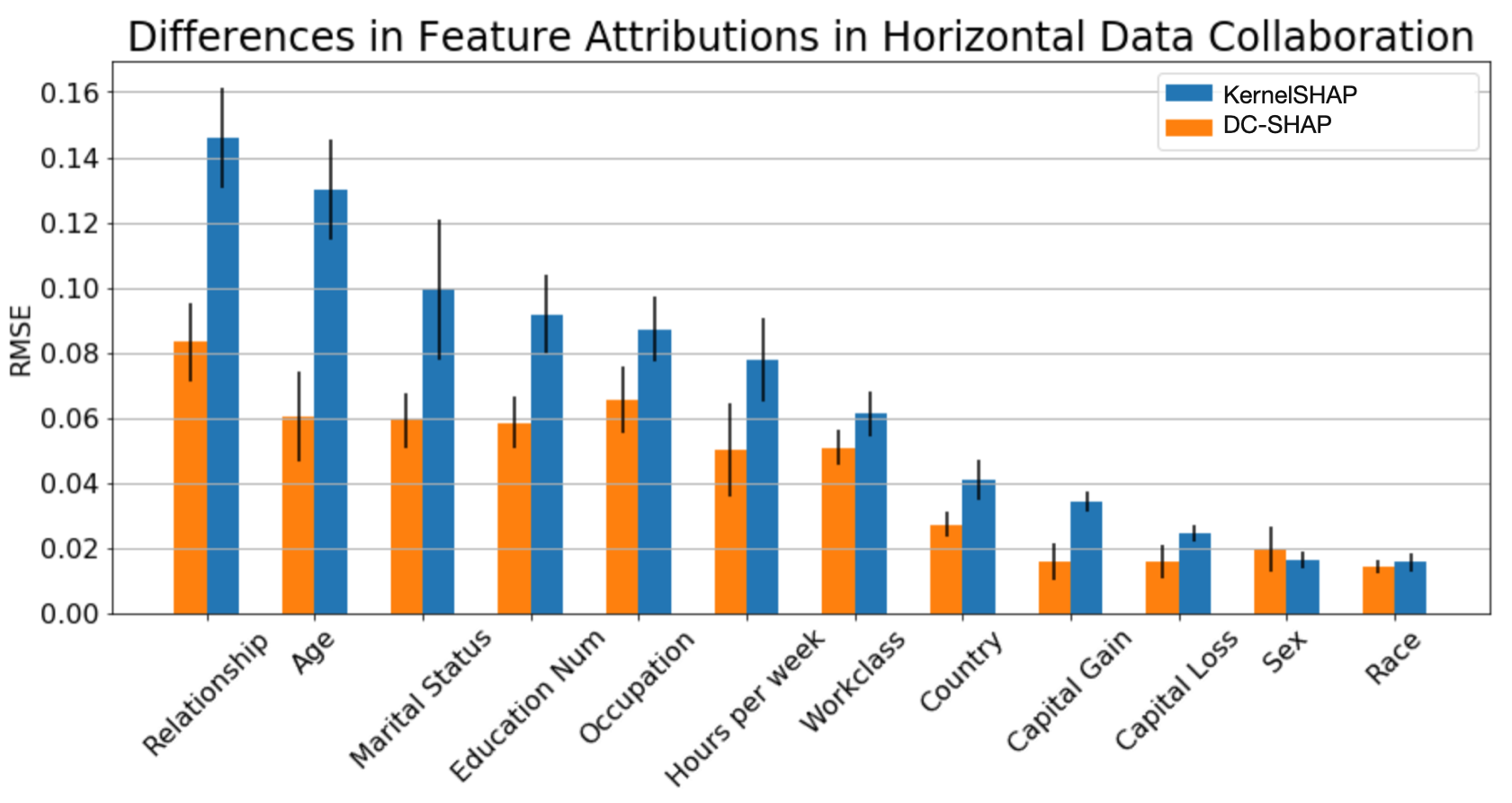}
\caption{Differences in feature attributions among the two users in horizontal Data Collaboration expressed as Root Mean Square Error for each feature. \label{fig:hor}}
\end{figure}

\subsection{Consistency of Explanations in Horizontal Collaboration}
For testing the consistency of the proposed feature attribution method in horizontal data collaboration, we split the training data among two users and trained a Data Collaboration model as described in Figure \ref{fig:dc1}. After that, we compared explanation coefficients between the two users for the same 50 samples of the test set and reported the difference in the attribution of each feature measured by Root Mean Square Error (RMSE). This metric was preferred over the similar Mean Absolute Error (MAE) metric because RMSE tends to give higher weight to large errors and is recommended when large errors are particularly undesirable, as it is in our case. First, the simulated users obtain explanations for the samples of the test set using KernelSHAP method as described in Algorithm \ref{appendix:shap}, computing the reference value from their share of the training data. Then, the explanations for the same samples are obtained with the proposed Horizontal DC-SHAP method as described in Figure \ref{fig:shap_dc1} and Algorithm \ref{appendix:hor}. The divergence of the explanations between the two users obtained by both methods is reported in Figure \ref{fig:hor}. The experiment was repeated 10 times, with random seeds set up from 0 to 9. 

We then repeated the same experiment on other open access datasets with similar learning tasks: Iris,  Wine, Heart Decease from UCI Machine Learning Repository \cite{Dua:2019}, and Pima Indian Diabetes \cite{smith1988using}. We did not change the hyper-parameters of the DC model apart from the target dimension for dimensionality reduction, which was adjusted to 3/4 of the original dimension of each dataset. Table \ref{T1} reports the results for each dataset averaged across all features. The results indicate better consistency in feature attribution among the users when the proposed method is used. In particular, the average discrepancy in feature attributions in terms of RMSE decreased in various datasets by at least a factor of 1.75.

\begin{table}[!h]
\centering
\caption{Differences in feature attributions among the two users in horizontal Data Collaboration expressed as Root Mean Square Error averaged across all features on various open datasets}

\begin{tabular}{cccccc}

\hline
Data & DC model & KernelSHAP & DC-SHAP\\
(\# features)&Accuracy&RMSE&RMSE\\
\hline
\texttt{Iris(4)} & 0.95 & 0.26\textpm{0.04} & 0.09\textpm{0.02}\\ 
\texttt{Pima(8)} & 0.73 & 0.07\textpm{0.01} & 0.01\textpm{0.00}\\ 
\texttt{Adult(12)} & 0.83 & 0.07\textpm{0.01} & 0.04\textpm{0.01}\\ 
\texttt{Wine(13)} & 0.94 & 0.10\textpm{0.03} & 0.02\textpm{0.00}\\ 
\texttt{Heart(13)} & 0.80 & 0.09\textpm{0.01} & 0.04\textpm{0.01}\\ 
\hline
\end{tabular}\\[9pt]

\label{T1}
\end{table}


\begin{figure}
\centering\includegraphics[width=1\linewidth]{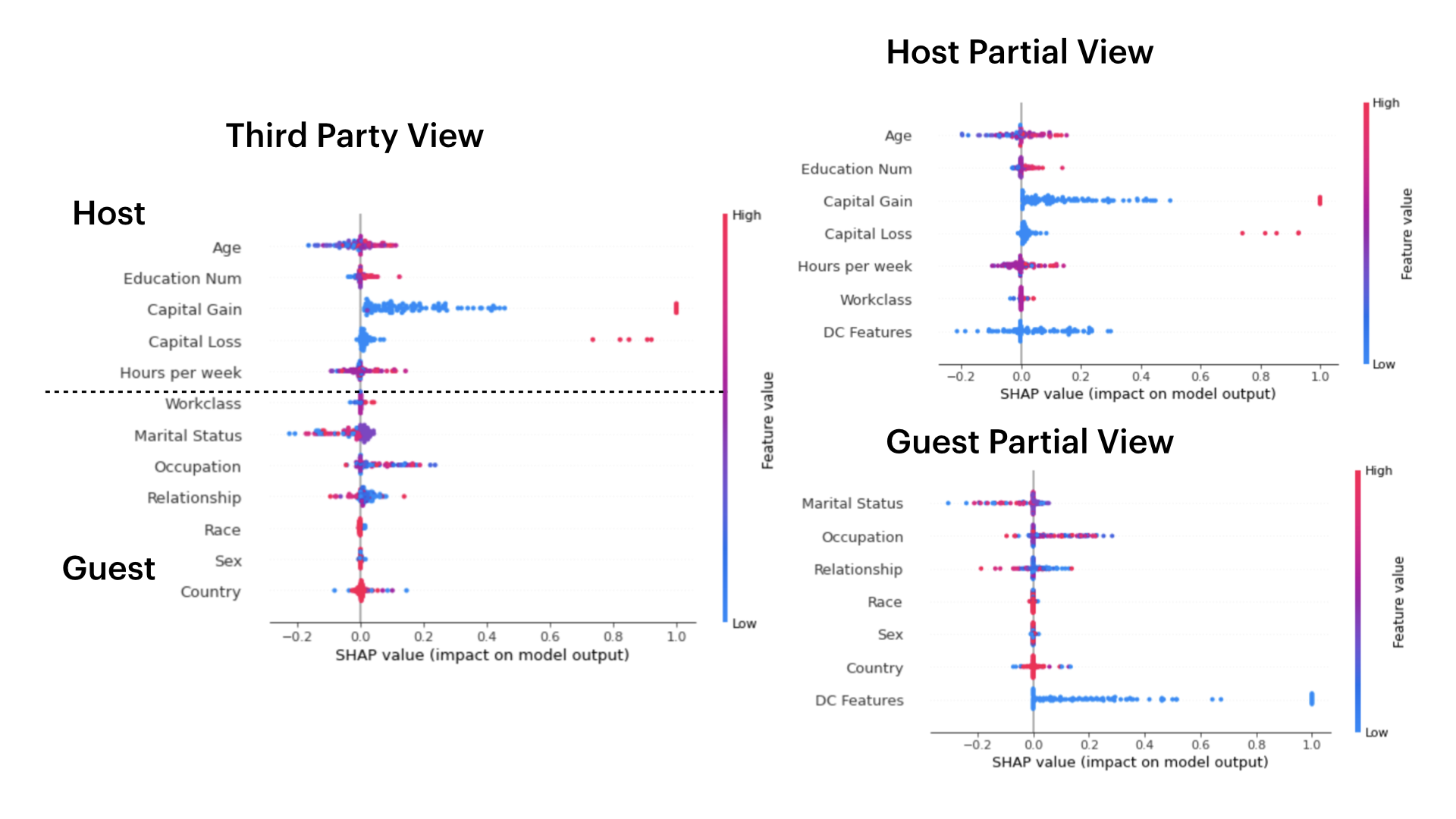}
\caption{Feature attribution comparison in Vertical Data Collaboration. Each dot represents a feature of one data instance. Features are listed in the fixed order to improve readability.\label{fig:vert}}
\end{figure}

\subsection{Explanations in Vertical Collaboration}
\begin{figure}

\centering\includegraphics[width=1\linewidth]{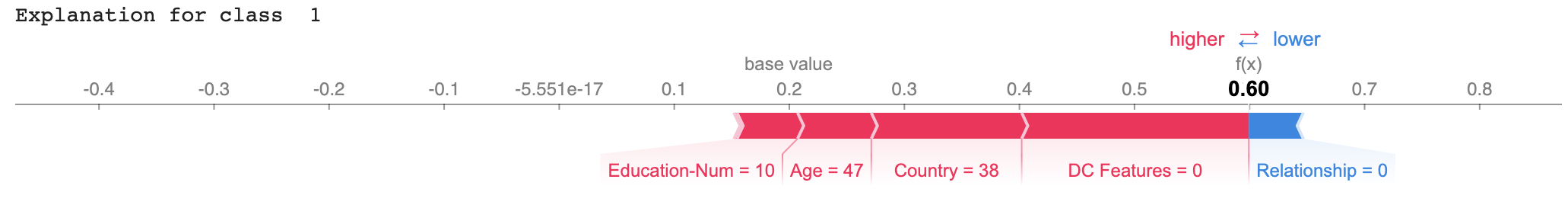}
\caption{Example of a single sample DC-SHAP output. The predicted value for this sample is 0.6 as class 1, and feature values in red have a positive impact on the model prediction while the feature in blue have a negative impact. \label{fig:example} }
\end{figure}

In Vertical DC, there is no problem with diverging baselines because collaborating parties share the same set of training samples. The challenge here is that parties have disjoint sets of features and in order to evaluate the model, they have to follow a protocol of information exchange that does not leak private data and allows to obtain valid explanations of model predictions. Algorithm \ref{appendix:vert_all} and \ref{appendix:vert_part}, achieve this by exchanging intermediate representations of vertically partitioned data inputs into explainability algorithm and by balancing partial feature attributions with a unified \textit{DC Features} indicator that reflects a combined effect of hidden features on model output (see example output in Figure \ref{fig:example}).    

To demonstrate the proposed feature attribution method in vertical Data Collaboration, we let host and guest parties share all rows of the Adult training data but with fully disjoint sets of features. In our experiments, the host was assigned features from 0 to 5, and the guest - from 6 to 11. With this partition, we trained a K-Nearest Neighbour classification model through Data Collaboration and obtained feature attributions for 100 test samples data.

The results in Figure~\ref{fig:vert} show feature attributions for all features as if requested by a third party and partial feature attribution results obtained by the host and the guest party, with external features being aggregated into one \textit{DC Features} indicator. It can be observed that partial results are consistent with the results for all features.

\section{Conclusion}

In this work, we identified and demonstrated the problems with explainability in distributed machine learning that can impede the transparency of AI-based decision-making systems. 
We further addressed those problems by proposing an Explainable Data Collaboration Framework based on model-agnostic additive feature attribution algorithm (KernelSHAP) and Data Collaboration method of privacy-preserving distributed machine learning. The proposed framework consists of three algorithms that tackle different use-case scenarios of  horizontal and vertical distributed machine learning, while ensuring that end-users and participants of the collaboration obtain consistent and accurate feature attributions for overlapping samples. The performance of the proposed algorithms was experimentally verified on several open-access datasets. Our method achieved a significant (by at least a factor of 1.75) decrease in feature attribution discrepancies among the users in horizontally partitioned machine learning, and proved capable of obtaining explanations in vertically partitioned machine learning, such that a partial feature view was consistent with a full feature view.


\subsection{Limitations and Future Work}
The presented method relies on the KernelSHAP algorithm, which is a brute-force method of obtaining Shapley values for any model. It is not suitable for high-dimensional data because it relies on constructing  a power set of all features, which quickly becomes intractable as the number of features increase. In the future work, we plan to adapt existing methods of approximating Shapley values as well as more optimized versions of SHAP for specific models, such as TreeExplainer and DeepExplainer.

\newpage
\appendix
\section{Appendix}

\begin{algorithm}
\caption{SHAP values calculation}
\label{appendix:shap}

\textbf{Input}\raggedright: $x$ instance of interest  \\
\textbf{Input}\raggedright: $f$, machine learning model \\
\textbf{Input}\raggedright: $r$, reference value \\
\textbf{Input}\raggedright: $M$, number of features  \\
\textbf{Output}\raggedright: SHAP values $\phi$ for all features of $x$\\

\begin{algorithmic}[1]

\STATE Construct power set of all features $S\in \mathbb{R}^{{2^M}\times{M}}$ \\
\STATE Make new data set $X'\in \mathbb{R}^{{2^M}\times{M}}$ \\
\STATE Make an empty array of weights $w \in \mathbb{R}^{2^M}$ \\

\FOR{$s$ in $S$}
\STATE Set $x'$ positions that are in $s$ to be original values of $x$ \\
\STATE Set $x'$ positions that are not in $s$ to be reference values of $r$\\
\STATE Calculate weights for current combination of features $w_s$ as ${M-1}/{({M \choose |s|})|s|(M-|s|)}$ \\ 
\ENDFOR \\

\STATE Get model predictions for the constructed dataset $y' = f(X')$ \\
\STATE Fit weighted linear model $l$ to $(S, w, y') $
\RETURN Coefficients of $l$ as SHAP values $\phi$
\end{algorithmic}
\end{algorithm}

\begin{algorithm}
\caption{Horizontal DC-SHAP for calculating SHAP values for the features of one user}
\label{appendix:hor}
\textbf{Input}\raggedright: $x \in \mathbb{R}^{1{\times}n}$, user features for the sample of interest  \\
\textbf{Input}\raggedright: $X_{anc} \in \mathbb{R}^{m_{anc}{\times}n}$, anchor data common among all users \\
\textbf{Input}\raggedright: $F$, individual dimensionality reduction function \\
\textbf{Input}\raggedright: $G$, individual integrating function \\
\textbf{Input}\raggedright: $h$, machine learning model \\
\textbf{Output}: SHAP values for all features\\
\begin{algorithmic}
\STATE Set reference value $r\in \mathbb{R}^{1{\times}n}$ to the median values of $X_{anc}$ \\
\RETURN Algorithm 1 with inputs $x$, $r$, $f=F(G(h))$, $M=n$ \\
\end{algorithmic}
\end{algorithm}

\begin{algorithm}
\caption{DC-SHAP for calculating SHAP values for all features in vertical Data Collaboration}
\label{appendix:vert_all}
\textbf{Input}\raggedright: $x \in \mathbb{R}^{1{\times}n}$, all features for the sample of interest  \\
\textbf{Input}\raggedright: $r \in \mathbb{R}^{1{\times}\hat{n}}$, reference values for all features \\
\textbf{Input}\raggedright: $F_i$, individual dimensionality reduction functions of all users\\
\textbf{Input}\raggedright: $f$, machine learning model \\
\textbf{Output}: SHAP values for all features\\
\begin{algorithmic}[1] 

\STATE Construct power set of all features $S\in \mathbb{R}^{{2^n}\times{n}}$ and corresponding weights array $w$\\
\STATE From $x$, $r$ and $S$ construct $X'$ as described in Algorithm \ref{appendix:shap} \\

\COMMENT{SERVER SIDE} \\
\STATE Split $X'$ vertically among the users so that each $X'^{j}$ has features corresponding to users' data 
\STATE For each user, apply function $F_i$ to $X'^j$ \\
\STATE Concatenate $X'^j$ into one dataset $\hat{X}$
\STATE Run model prediction $h(\hat{X})$ obtaining labels $y'$\\
\COMMENT{USER SIDE} \\
\STATE Calculate SHAP values $\phi(x')$ with $S$, $w$, and $y'$ as described in Algorithm \ref{appendix:shap}\\
\RETURN $\phi(x')$ \\

\end{algorithmic}
\end{algorithm}

\begin{algorithm}
\caption{DC-SHAP for calculating partial SHAP values in vertical DC}
\label{appendix:vert_part}
\textbf{Input}\raggedright: $x_h \in \mathbb{R}^{1{\times}h}$, host features for the input of interest  \\
\textbf{Input}\raggedright: $\hat{x}_g \in \mathbb{R}^{1{\times}\hat{g}}$, guest features for the input of interest\\
\textbf{Input}\raggedright: $r_h,\in \mathbb{R}^{1{\times}h}$, reference values for host features\\
\textbf{Input}\raggedright: $\hat{r}_g,\in \mathbb{R}^{1{\times}\hat{g}}$, reference values for guest features \\
\textbf{Input}\raggedright: $F_h$, dimensionality reduction function of the host\\
\textbf{Input}\raggedright: $h$, machine learning model \\
\textbf{Output}: SHAP values for host features and aggregated guest feature\\

\begin{algorithmic}[1] 
\STATE Set $M$, number of features for attribution to $h+1$
\STATE Construct power set $S\in \mathbb{R}^{{2^M}\times{M}}$ and weights array $w$\\
\STATE Make new data set $X'_h\in \mathbb{R}^{{2^M}\times{h}}$ \\
\STATE Make new data set $X'_g\in \mathbb{R}^{{2^M}\times{\hat{g}}}$ \\
\FOR{$s$ in $S$}
\STATE Set $x'_h$ positions that are in $s$ to be original values of $x_h$ \\
\STATE Set $x'_h$ positions that are not in $s$ to be reference values $r_h$\\
\IF{$s[-1]$} 
\STATE Set $x'_g$ to $\hat{x}_g$ 
\ELSE 
\STATE Set $x'_g$ to reference value $\hat{r}_g$
\ENDIF
\ENDFOR
\STATE Transform $X'_h$ with function $F_h$
\STATE Get model predictions $y'$ for concatenated data sets [$F_h(X'_h), X'_g$]
\STATE Calculate SHAP values $\phi(x')$ from $S$, $w$, and $y'$ with Algorithm \ref{appendix:shap} \\
\RETURN $\phi(x')$

\end{algorithmic}
\end{algorithm}

\newpage

\bibliographystyle{splncs04}
\bibliography{mybib}






\end{document}